\documentclass[conference]{IEEEtran}
\pdfoutput=1
\IEEEoverridecommandlockouts
\usepackage{cite}
\usepackage{amsmath,amssymb,amsfonts,thmtools}
\usepackage{algorithmic}
\usepackage{textcomp}
\def\BibTeX{{\rm B\kern-.05em{\sc i\kern-.025em b}\kern-.08em
    T\kern-.1667em\lower.7ex\hbox{E}\kern-.125emX}}

\usepackage{graphicx,amsmath,amsfonts,amssymb,hyperref,float}
\usepackage{tabularray}
\usepackage[dvipsnames]{xcolor}
\usepackage{tikz}
\usetikzlibrary{arrows.meta,arrows,hobby}
\tikzset{
    >={Straight Barb[scale=0.8]}
}
\usepackage{tikz-3dplot}
\usepackage{pdfpages}

\ifCLASSOPTIONcompsoc
    \usepackage[caption=false, font=normalsize, labelfont=sf, textfont=sf]{subfig}
\else
\usepackage[caption=false, font=footnotesize]{subfig}
\fi

\tdplotsetmaincoords{80}{0}
\usetikzlibrary{calc}

\declaretheorem[numberwithin=section]{theorem}
\declaretheorem[sibling=theorem]{proposition}

\DeclareMathOperator{\GL}{GL}
\DeclareMathOperator{\SO}{SO}
\DeclareMathOperator{\SE}{SE}
\DeclareMathOperator{\Ad}{Ad}
\DeclareMathOperator{\ad}{ad}

\newcommand{\lie}[1]{\mathfrak{#1}}
\newcommand{\mfd}[1]{\mathcal{#1}}
\newcommand{\mat}[1]{\boldsymbol{#1}}
\newcommand{\norm}[1]{\left\Vert#1\right\Vert}
\newcommand{\data}[1]{\mbox{\textit{#1}}}

\begin{document}

\title{Equivariant Filter for Tightly Coupled LiDAR-Inertial Odometry*\\
\thanks{%
  \textsuperscript{1}Anbo Tao, Chunxi Xia, and Xingxing Li are with the School of 
  Geodesy and Geomatics, Wuhan University, China.
  \{tao\_ab,xiachunxi\}@whu.edu.cn, xxli@sgg.whu.edu.cn
}
\thanks{%
  \textsuperscript{2}Yarong Luo and Chi Guo are with 
  the GNSS Research Center, Wuhan University, China.
  \{yarongluo,guochi\}@whu.edu.cn
}
\thanks{%
  *This work is supported in part by
  the National Natural Science Foundation of China under
  Grant 42404025,
  the China Postdoctoral Science Foundation 
  under Grant Number(2023TQ0248) 
  and the National Key Research and Development 
  Program of China under Grant 2021YFB2501102 and 2023YFB3907100
  (Corresponding author: Yarong Luo).
}
}

\author{
  \IEEEauthorblockN{%
    Anbo Tao\textsuperscript{1},
    Yarong Luo\textsuperscript{2},
    Chunxi Xia\textsuperscript{1},
    Chi Guo\textsuperscript{2}, and
    Xingxing Li\textsuperscript{1}
  }
}

\maketitle

\begin{abstract}

  Pose estimation is a crucial problem in simultaneous localization and mapping (SLAM).
  However, developing a robust 
  and consistent state estimator remains a significant challenge, as the 
  traditional extended Kalman filter (EKF) struggles to handle the model
  nonlinearity, especially for inertial measurement unit (IMU) and 
  light detection and ranging (LiDAR). To provide a consistent and efficient
  solution of pose estimation, we propose Eq-LIO, a 
  robust state estimator for tightly coupled LIO systems based on an 
  equivariant filter (EqF). Compared with the invariant 
  Kalman filter based on the $\SE_2(3)$ group structure, the EqF uses the symmetry 
  of the semi-direct product group to couple the system state including IMU bias, 
  navigation state, and LiDAR extrinsic calibration state, thereby suppressing 
  linearization error and improving the behavior of the estimator 
  in the event of unexpected state changes. The proposed Eq-LIO owns natural 
  consistency and higher robustness, which is theoretically proven with 
  mathematical derivation and experimentally verified through a series of tests on
  both public and private datasets.
\end{abstract}


\section{Introduction}


In recent years, light detection and ranging (LiDAR) sensors have gained widespread use in 
simultaneous localization and mapping (SLAM) due to their ability to 
capture precise depth information. However, LiDAR is vulnerable to 
distortion caused by rapid movements. In contrast, inertial 
measurement units (IMUs) can provide motion data at high sampling rates 
regardless of external environmental conditions. 
IMU and LiDAR exhibit complementary characteristics, integrating 
these sensors can significantly enhance the continuity and accuracy of 
the output \cite{liom,lio-sam,qin2020lins}.


As a result, LiDAR-inertial odometry (LIO) has been widely used in
industry. Among various approaches, filter-based methods 
are well-suited for low-cost platforms with limited 
computing resources, as they can efficiently handle large volumes 
of real-time measurements while maintaining lower computational demands.
Among filter-based methods, the extended Kalman filter (EKF) achieves 
high efficiency, albeit with some loss of accuracy. This trade-off 
makes it a classic data fusion method, as seen in applications like 
\cite{aghili2016robust} and others.
However, researchers have identified that EKF may yield 
inconsistent estimates, where the computed covariance deviates from the 
true covariance \cite{huang2007convergence}. 
To alleviate the inconsistency problem, Barrau
and Bonnabel proposed the invariant extended Kalman filter (IEKF) by analyzing the system's group affine properties
\cite{IEKF1}.
Later, Shi et al. \cite{shi2023invariant} applied the imperfect-IEKF for fusing
IMU and LiDAR measurements to reduce estimation error, but the IMU bias will
destroy the symmetry of the system. 
Mahony et al. proposed a modern method known as equivariant filtering (EqF)
\cite{EqF1,10179117}, which naturally enhances consistency and 
robustness with lower computational overhead by incorporating 
different system symmetries and fixed linearization points. 
In light of these advantages, our work fully leverages EqF.


In this article, we introduce Eq-LIO, a fast and reliable tightly coupled LIO 
framework based on the EqF state estimator. Our approach leverages the 
semi-direct product group to incorporate symmetries, including bias 
and extrinsic parameters, into the LIO framework. Additionally, we optimize 
the gravity constraint on the manifold $\mathbb{S}^2$. 
The error dynamics of the EqF ensures that it always performs 
linearization operations at a fixed origin, which is the 
key to improving system consistency and reducing linearization errors.
Finally, we validate the performance of Eq-LIO in both standard and challenging scenarios.
The main contributions of our work are as follows:
\begin{enumerate}
  \item We introduce Eq-LIO, a tightly coupled LIO system based on 
  the EqF state estimator, with LiDAR and IMU self-calibration
  capabilities and gravity constraints on $\mathbb{S}^2$.
  To the best of our knowledge, this is the first LIO system to utilize an 
  equivariant filter.
  \item We perform extensive benchmark testing across diverse datasets, 
  demonstrating that the proposed framework delivers superior accuracy, 
  robustness, and consistency without increasing computational demands.
  To support community development, our source code is available online:
  \url{https://github.com/Eliaul/Eq-LIO}.
\end{enumerate}

\section{Related Works}

In this section, we focus on the most related works on LIO and Kalman filtering.

\subsection{LiDAR-Inertial Odometry}


LIO methods can be broadly categorized into two approaches: 
loosely coupled methods and tightly coupled methods. Loosely coupled methods typically 
process measurements from the IMU and LiDAR independently before 
fusing the results. For instance, the LOAM algorithm proposed in 
\cite{loam} utilized the pose estimated from IMU data as an initial 
guess for LiDAR scan registration. In \cite{GPS/INS}, an adaptive EKF 
was employed to fuse IMU measurements with poses estimated by LiDAR SLAM, 
enhancing the accuracy of pose estimation. The direct LiDAR odometry method 
introduced in \cite{DLO} enabled real-time, high-speed, and 
high-precision processing of dense point clouds.
While loosely coupled methods offer high computational 
efficiency, their limited ability to effectively fuse sensor 
data may lead to unreliable results in the later stages of 
the fusion process.


The tightly coupled methods directly fuse LiDAR point cloud 
data with IMU measurements to obtain an optimal state estimation,
which can be divided into two primary categories: optimization-based methods 
and filter-based methods. For optimization-based methods, Ye et al.\cite{liom} introduced 
the LIOM framework, which applied sliding window optimization to LIO. 
Similarly, LIO-SAM \cite{lio-sam} employed smoothing and mapping 
techniques to achieve sensor fusion and global optimization.
In filter-based methods, LINS \cite{qin2020lins} utilized a robo-centric iterative Kalman filter for
tightly coupled pose optimization. 
FAST-LIO \cite{fast-lio} utilized a world-centric iterative Kalman filtering 
for state estimation, incorporating a novel Kalman gain formula to 
enhance computational efficiency. FAST-LIO2 \cite{fastlio2} further improved 
system efficiency and robustness by using direct point cloud 
registration and an ikd-tree structure.

\subsection{Kalman Filtering for Navigation Applications}


Since the introduction of the classic extended Kalman filter (EKF), 
it has become a widely used tool in filter-based SLAM algorithms. 
However, the traditional EKF-based LIO algorithm is susceptible to overconfidence 
issues due to spurious information gain in unobservable directions.
To address the 
inconsistency issues inherent in EKF, Barrau and Bonnabel proposed the 
invariant extended Kalman filter (IEKF) \cite{IEKF1}, demonstrating 
that a consistent filtering algorithm can be developed on group affine systems.
This method assumes that the IMU is bias-free, which could not be 
ignored in practical applications.
Despite this limitation, 
many researchers \cite{IEKF-exa1}, \cite{IEKF-exa2}  continue 
to use this imperfect IEKF framework in the design of SLAM algorithms.
They defined the IMU bias on $\mathbb{R}^6$, which destroys 
the group affine property of the system.
Recently, SuIn-LIO \cite{SuIn-LIO} was proposed, combining an 
IEKF with an efficient surfel-based map to achieve 
high-precision LIO.
Shi et al.\cite{shi2023invariant} applied the IEKF to the 
robo-centric and world-centric based LIO and demonstrated 
that the world-centric method achieves a higher level of accuracy.



Unlike IEKF, which parameterizes the state using Lie groups, Mahony et al. 
developed an equivariant filtering (EqF) framework \cite{EqF1}, \cite{EqF2} that generalizes 
the state space from Lie groups to homogeneous spaces.
Additionally, EqF does not require constant 
changes to the linearization point, which naturally ensures system consistency.
Subsequently, Fornasier et al. applied EqF to an inertial navigation system \cite{EqF4}, 
which accounts for bias, and demonstrated that various filters, 
including the EKF, can be incorporated within the EqF framework \cite{EqF3}.
Recently, Fornasier et al. \cite{MSCEqF} introduced an equivariant 
filter design for vision-assisted INS systems. This new design has 
been demonstrated to outperform state-of-the-art methods in both 
robustness and transient behavior.

\section{Mathematical Preliminaries and Notation}

\subsection{Smooth Manifolds and Lie Theory}

Let $\mfd{M}$ be a smooth manifold, we use $T_\xi \mfd{M}$ to denote the tangent space
of $\mfd{M}$ at point $\xi\in \mfd{M}$ and $T\mfd M$ to denote the tangent bundle. The notion $\mathfrak{X}(\mfd{M})$
denotes the set of smooth vector field on $\mfd{M}$, where each element of $\mathfrak{X}(\mfd{M})$ assigns 
a tangent vector in $T_\xi \mfd{M}$ to each point $\xi\in \mfd{M}$. A Lie group $G$ is a smooth manifold with a
group structure, and group multiplication and inversion operations are smooth as mappings between manifolds.
For any elements $X,Y\in G$, the group multiplication is denoted by $XY$, the inverse of $X$
is denoted by $X^{-1}$, the identity is denoted by $I$. The tangent space of $G$ at $I$ is called 
the Lie algebra of $G$, denoted by $\lie g$, that is, $\lie g=T_IG$. The Lie algebra $\lie g$ is a
vector space with dimension equal to $n=\dim G$. So we have an isomorphism from $\lie g$
to $\mathbb{R}^n$, which we denote $(\cdot)^\vee:\lie g\to\mathbb{R}^n$, and the inverse of
$(\cdot)^\vee$ is $(\cdot)^\wedge:\mathbb{R}^n\to\lie g$.
In this work, 
we focus only on the Lie group of matrices, i.e., $G$ is a subgroup of the general linear
group $\GL_n(\mathbb{R})$.

Fixing a point $X\in G$, there are two important mappings $L_X:G\to G$ and 
$R_X:G\to G$, which are called left translation and right translation respectively:
\[                      
  L_X(Y)=XY,\quad R_X(Y)=YX.
\]

Given an element $X\in G$, consider the conjugate action $Y\mapsto XYX^{-1}$ on 
the Lie group. The differential of this action at the identity $I$ is
\[
  \Ad_X:\lie g\to \lie g,\quad \Ad_X(\mat u)= (dL_X)\circ (dR_{X^{-1}})(\mat u),
\]
for every $\mat u\in\lie g$. This map $\Ad_X$ is called a (big) Adjoint map. 
Since we have an isomorphism $(\cdot)^\vee:\lie g\to\mathbb{R}^n$, 
$\Ad_X^\vee := (\cdot)^\vee\circ \Ad_X\circ(\cdot)^\wedge$, 
as a linear map $\mathbb{R}^n\to\mathbb{R}^n$, can be viewed as a matrix.

For any $u\in\lie g$, the (little) adjoint map is defined by
\[
  \ad_u:\lie g\to\lie g,\quad \ad_u(v)=[u,v],  
\]
where $[u,v]$ is the Lie bracket.

\subsection{Group Action, Useful Maps and Notation Explanation}

Assuming $G$ is a Lie group and $\mfd M$ is a smooth manifold, we consider the Lie group action $\phi:G\times\mfd M\to\mfd M$. 
Fixing an element $X\in G$, we write $\phi_X:\mfd M\to\mfd M$ to represent
\[
  \phi_X(\xi):=\phi(X,\xi)  .
\]
Fix an element $\xi\in\mfd M$, we write $\phi^{(\xi)}:G\to\mfd M$ to represent
\[
  \phi^{(\xi)}(X):=\phi(X,\xi).  
\]

For all $\mat X=(\mat C,\mat a,\mat b)\in\SE_2(3)=\SO(3)\ltimes (\mathbb{R}^3\oplus\mathbb{R}^3)$,
define $\Gamma$ to extract the rotation and the position part of $\mat X$:
\[
  \Gamma:\SE_2(3)\to\SE(3),\quad \Gamma(\mat X)=(\mat C,\mat b).  
\]

In this article, the use of bold and non-bold letters for 
the same character represents two different meanings. 
Bold letters are generally used to denote matrices, 
while non-bold letters are typically used to represent elements 
in groups or manifolds. The proofs of this paper are all 
available in the supplementary material\cite{supp}.

\section{System Overview}\label{sec:system}

\begin{figure}[tb]
  \centering
  \includegraphics[width=\linewidth]{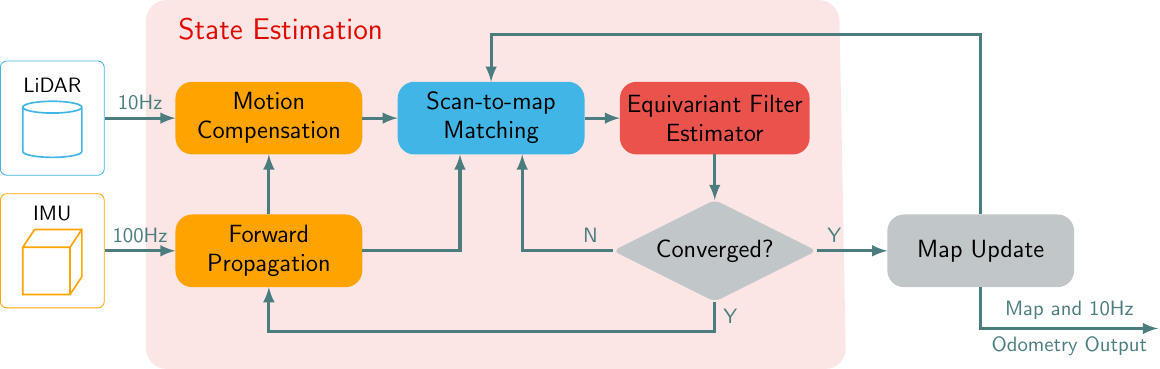}
  \caption{%
    Overview of Eq-LIO.
    The equivariant filtering state estimator is employed in the 
    odometry module, with the measurement model constructed through 
    scan-to-map matching.
    The red box in the figure highlights the main content of this paper. 
    For detailed information, please refer to Section \ref{sec:system} and \ref{sec:eqf}.
  }\label{fig:flow}
\end{figure}

The pipeline of the proposed Eq-LIO is illustrated in Fig. \ref{fig:flow}.
The raw point cloud is first de-skewed using IMU predicted pose \cite{fastlio2}.
Next, a world-centric equivariant 
filter state estimator is constructed for the odometry module, 
utilizing the scan-to-map matching method to establish correspondence 
between the LiDAR scan and the global map. Finally, the output 
from the state estimator is used to update the global map.

\subsection{System Definition}

The IMU coordinate system at the origin is used as the reference coordinate system, 
which we call the world coordinate system and is represented by $w$.
We use $b$ to represent the body coordinate system.
The system model of LIO is as follows.
\begin{equation}\label{eq:primitive dynamic model}
  \begin{aligned}
    \dot{\mat C}_{b}^w &= \mat C_{b}^w(\mat\omega-\mat b_g)^\wedge,
    \\
    \dot{\mat v}_{wb}^w &= \mat C_{b}^w(\mat a-\mat b_a)+\mat g^w,
    \\
    \dot{\mat r}_{wb}^w &= \mat v_{wb}^w, 
    \\
    \qquad \dot{\mat b}_g &= \mat 0_{3\times 1},\quad
    \dot{\mat b}_a = \mat 0_{3\times 1},\\
    \dot{\mat C}^b &= \mat O_{3\times 3},\quad 
    \dot{\mat l}^b = \mat 0_{3\times 1},
  \end{aligned}
\end{equation}
where $\mat C_b^w$, $\mat v_{wb}^w$, $\mat r_{wb}^w$ represents the attitude, velocity and position 
of the IMU, $\mat b_g$ and $\mat b_a$ represent the gyroscope bias and 
accelerometer bias, respectively, and $\mat C^b$ and $\mat l^b$ represent 
the attitude and position of the LiDAR relative to the IMU, $\mat g^w$ 
represents the gravity vector, $\mat \omega$ and $\mat a$ represent the 
angular velocity and specific force measured by the IMU.

We introduce a virtual ``velocity bias" $\mat b_\mu$ and virtual inputs 
$\mat\mu,\mat\tau_c,\mat\tau_g,\mat\tau_a,\mat\tau_\mu,\mat\tau_l$ to exploit the geometric properties of 
the system, the equation \eqref{eq:primitive dynamic model} expands to \cite{EqF4}:
\begin{equation}\label{eq:virtual dynamic model}
  \begin{alignedat}{2}
    \dot{\mat C}_{b}^w &= \mat C_{b}^w(\mat\omega-\mat b_g)^\wedge
      ,
      &\qquad \dot{\mat b}_g &= \mat \tau_g,
    \\
    \dot{\mat v}_{wb}^w &= \mat C_{b}^w(\mat a-\mat b_a)
      +\mat g^w, 
      & \dot{\mat b}_a &= \mat \tau_a,
    \\
    \dot{\mat r}_{wb}^w &= \mat C_b^w(\mat \mu-\mat b_\mu)
     +\mat v_{wb}^w, 
      & \dot{\mat b}_\mu &= \mat \tau_\mu,
    \\
    \dot{\mat C}^b &= \mat C^b\mat \tau_c^\wedge,
      & \dot{\mat l}^b &= \mat \tau_l.
  \end{alignedat}
\end{equation}


The system state is modeled as $\xi=(\mat T,\mat b^\wedge,\mat K)\in\mfd M$,
with navigation state $\mat T=(\mat C_b^w,\mat v_{wb}^w,\mat r_{wb}^w)\in\SE_2(3)$,
bias $\mat b=(\mat b_g,\mat b_a,\mat b_\mu)\in\mathbb{R}^9$ and 
extrinsic parameters $\mat K=(\mat C^b,\mat l^b)\in\SE(3)$ of LiDAR and IMU. 
Let $\mat u=(\mat w^\wedge,\mat g^\wedge,\mat\tau^\wedge,\mat\tau_k^\wedge)\in(\lie{se}_2(3))^3\times\lie{se}(3)$ 
be the system inputs, where 
$\mat g^\wedge=(\mat 0^\wedge,\mat g^w,\mat 0)\in\lie{se}_2(3)$, 
$\mat \tau^\wedge=(\mat\tau_g^\wedge,\mat\tau_a,\mat\tau_u)\in\lie{se}_2(3)$,
$\mat\tau_k^\wedge=(\mat\tau_c^\wedge,\mat\tau_l)\in\lie{se}(3)$ and 
$\mat w=(\mat\omega,\mat a,\mat\mu)$ be the IMU measurements.
In general, the system state evolves on the manifold $\mfd M=\SE_2(3)\times\lie{se}_2(3)\times \SE(3)$, 
and the system inputs are in the input space $L=(\lie{se}_2(3))^3\times\lie{se}(3)$.
In this notation, the system model \eqref{eq:virtual dynamic model} can be rewritten as
\begin{equation}\label{eq:dynamic model on manifold1}
  \begin{aligned}
    \dot{\mat T} &= f_1^0(\mat T)\mat T+\mat T(\mat w^\wedge-\mat b^\wedge)+\mat g^\wedge
    \mat T,\\
    \dot{\mat b} &= \mat\tau,\\
    \dot{\mat K} &=\mat K\mat\tau_k^\wedge, 
  \end{aligned}
\end{equation}
where $f_1^0:\SE_2(3)\to T\SE_2(3)$ defined by 
$\mat X=(\mat C,\mat v,\mat r)\mapsto (\mat O,\mat 0,\mat v)\in T_{\mat X}\SE_2(3)$
is a vector field on $\SE_2(3)$. 
We can write \eqref{eq:dynamic model on manifold1} in a more compact form:
\begin{equation}\label{eq:system equation}
  \begin{aligned}
    \dot{\xi}&=f^0(\xi)+f_{\mat u}(\xi) \\
    &=f^0(\xi)+\bigl(\mat T(\mat w-\mat b)^\wedge+\mat g^\wedge\mat T,\mat\tau^\wedge,\mat K\mat\tau_k^\wedge\bigr),
  \end{aligned}
\end{equation}
where $f^0:\mfd M\to T\mfd M$ defined by
$\xi\mapsto (f_1^0(\mat T),\mat O,\mat O)\in T_\xi\mfd M$
is a vector field on $\mfd M$.

\subsection{Equivariant Symmetry of the System}

The symmetry of the system arises from the key semidirect product group 
$G=\bigl(\SE_2(3)\ltimes\lie{se}_2(3)\bigr)\times\SE(3)$,
which is first introduced by \cite{EqF4}.
Let $X=(A,a^\wedge,B)\in G$, $\gamma=(\gamma_1^\wedge,\gamma_2^\wedge,\gamma_3^\wedge,\gamma_4^\wedge)\in L$.

\begin{proposition}\label{prop:phi}
  Define $\phi:G\times\mfd M\to\mfd M$ as
  \begin{equation}\label{eq:phi}
    \phi(X,\xi)=(\mat TA,\Ad_{A^{-1}}(\mat b^\wedge-a^\wedge),\Gamma(A)^{-1}\mat K B),
  \end{equation}
  where $\Ad_{A^{-1}}:\lie{se}_2(3)\to\lie{se}_2(3)$ represents the isomorphism induced by
  conjugation $Y\mapsto A^{-1}YA$ on $\SE_2(3)$. 
  Then $\phi$ is a transitive and free right group action of $G$ on $\mfd M$
  \cite{fornasier2022overcoming}.
\end{proposition}

\begin{proposition}\label{prop:psi}
  Define $\psi:G\times L\to L$ as 
  \begin{multline}
    \psi(X,\gamma)=\bigl(\Ad_{A^{-1}}(\gamma_1^\wedge-a^\wedge)+f_1^0(A^{-1}),\\
    \gamma_2^\wedge,\Ad_{A^{-1}}\gamma_3^\wedge,\Ad_{B^{-1}} \gamma_4^\wedge\bigr),
  \end{multline}
  then $\psi$ is a right group action of $G$ on $L$.
\end{proposition}

\begin{theorem}
  System \eqref{eq:system equation} is equivariant with respect to group action in \autoref{prop:phi}
  and \autoref{prop:psi}, that is,
  \[
    f^0(\xi)+f_{\psi_X(\gamma)}(\xi)
    =d\phi_X\bigl(f^0(\phi_{X^{-1}}(\xi))+f_\gamma(\phi_{X^{-1}}(\xi))\bigr)
  \]
  for every $X\in G,\xi\in\mfd M$ and $\gamma\in L$.
\end{theorem}

\subsection{Lifted System}

Equivariant filtering requires a lift $\Lambda:\mfd M\times L\to\lie g$ to transfer the system differential equations 
to the Lie group $G$. The lift $\Lambda$ requires
\[
  d\phi^{(\xi)}\circ\Lambda(\xi,\gamma)=f_\gamma(\xi),
\] 
where $\xi\in\mfd M$ and $\gamma\in L$.
If the group action $\phi$ is transitive,
then such a lift always exists.

\begin{theorem}
  Define $\Lambda_1:\mfd M\times L\to\lie{se}_2(3)$ as
  \[
    \Lambda_1(\xi,\gamma)=\gamma_1^\wedge-\mat b^\wedge+
    \Ad_{\mat T^{-1}}\gamma_2^\wedge+\mat T^{-1}f_1^0(\mat T),
  \]
  $\Lambda_2:\mfd M\times L\to \lie{se}_2(3)$ as
  \[
    \Lambda_2(\xi,\gamma)=\ad_{\mat b^\wedge}(\Lambda_1(\xi,\gamma))
    -\gamma_3^\wedge,  
  \]
  $\Lambda_3:\mfd M\times L\to\lie{se}(3)$ as
  \[
    \Lambda_3(\xi,\gamma)=\Ad_{\mat K^{-1}}\bigl[\Gamma(\Lambda_1(\xi,\gamma))\bigr]
    +\gamma_4^\wedge,
  \]
  then $\Lambda:\mfd M\times L\to \lie g=\lie{se}_2(3)\oplus\lie{se}_2(3)\oplus\lie{se}(3)$
  \begin{equation}\label{eq:Lambda}
    \Lambda(\xi,\gamma)=(\Lambda_1(\xi,\gamma),\Lambda_2(\xi,\gamma),\Lambda_3(\xi,\gamma))  
  \end{equation}
  is an equivariant lift \cite{EqF4}.
\end{theorem}

Let $X\in G$ be the state on the Lie group, and $\xi^0\in\mfd M$
be the original state.
If the group action $\phi$ is free, then the lift $\Lambda$ leads to
a system on Lie group $G$:
\begin{equation}
  \dot X=dL_X\circ\Lambda(\phi^{(\xi^0)}(X),\gamma).
\end{equation}
This transformation moves the estimation problem from the manifold to the Lie group, 
allowing the error to be defined via multiplication on the Lie group.

If the estimated error covariance reflects the true 
distribution of the errors, then the estimator is said to 
be consistent.
The consistency of the system can be proven using the group 
action $\phi$ in equation \eqref{eq:phi} and the lift $\Lambda$ in 
equation \eqref{eq:Lambda} \cite{van2021equivariant,wu2017invariant}.


\section{Equivariant Filter}\label{sec:eqf}

\subsection{Error Dynamics}

Fix the origin state $\xi^0=\bigl(\mat I_{5\times 5},\mat O_{9\times 9},\mat I_{4\times 4}\bigr)$.
For current state $\xi\in\mfd M$, define $e=\phi_{\hat{X}^{-1}}(\xi)=\phi_{X\hat X^{-1}}(\xi^0)=\phi_E(\xi^0)\in\mfd M$ 
to represent the error on $\mfd M$,
where $E=X\hat X^{-1}$ is right invariant error on $G$ \cite{mahony2022observer}.
Choose a local coordinate map $\varphi:U\to\mathbb{R}^9\oplus\mathbb{R}^9\oplus\mathbb{R}^6$ 
at the origin $\xi^0$ , $\varphi$ can be taken as
{\abovedisplayskip=4pt \belowdisplayskip=4pt
\[
  \varphi(e)=\log_G\circ\bigl( \phi^{(\xi^0)}\bigr)^{-1}(e)  .
\]}
Let $\varepsilon=\varphi(e)\in\mathbb{R}^{24}$.
The linear error $\varepsilon$ satisfies
\begin{align*}
  \dot\varepsilon&\approx \mat F\varepsilon+\mat B\mat n,\\
  \mat F&=(d\varphi)_{\xi^{(0)}}\circ \bigl(d\phi^{(\xi^0)}\bigr)_{I}
  \circ\bigl(d\Lambda_{\mat u}\bigr)_{\xi^0}\circ\bigl(d\varphi^{-1}\bigr)_{\mat 0},
\end{align*}
\vskip-3pt
\noindent
where $\mat n$ denotes the Gaussian noise. The state matrix $\mat F$ is
\begin{equation}
  \mat F=\begin{bmatrix}
    \mat F_{T} & \mat I_{9\times 9} & \mat O_{9\times 6}\\
    \mat O_{9\times 9} & \mat F_{b} & \mat O_{9\times 6} \\
    \mat F_{KT} & \mat F_{Kb} & \mat F_{K}
  \end{bmatrix}\in\mathbb{R}^{24\times 24},
\end{equation}
where
\begin{align*}
  \mat F_{T}&=\begin{bmatrix}
       \\
      {\mat g^w}^\wedge &  \\
      & \mat I_{3\times 3} & 
    \end{bmatrix},
  \mat F_{b}=\ad_{\Lambda_1(\xi^0,\mat u^0)}^\vee\in \mathbb{R}^{9\times 9},\\
  \mat F_{Kb}&=\begin{bmatrix}
    \mat I_{3\times 3} & \mat O_{3\times 3} & \mat O_{3\times 3}\\
    \mat O_{3\times 3} & \mat O_{3\times 3}  & \mat I_{3\times 3}
  \end{bmatrix}\in \mathbb{R}^{6\times 9},\\
  \mat F_{KT}&=\bigl(\mat I_{6\times 6}-\ad_{\Gamma(\Lambda_1(\xi^0,\mat u^0))}^\vee\bigr)
  \begin{bmatrix}
    \\
   & \mat I_{3\times 3} & 
 \end{bmatrix}\in \mathbb{R}^{6\times 9},\\
 \mat F_K&=\ad_{\Gamma(\Lambda(\xi^0,\mat u^0))}^\vee 
 \in \mathbb{R}^{6\times 6}.
\end{align*}
\vskip-3pt
\noindent
The letters with hat in them represent the estimated value of that state
and $\mat I$ represents the identity matrix. 
The derivations of the state matrix $\mat F$ and the input matrix $\mat B$
are privided in the supplementary material \cite{supp}.


\subsection{Measurement Model}

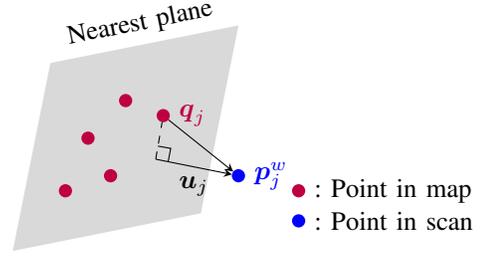
\begin{figure}[t]
  \centering
  \begin{tikzpicture}[
    scale=1,
    points/.style={
      circle,
      inner sep=0pt,
      outer sep=0pt,
      minimum width=5pt,
      fill=#1,
    }
  ]
    \fill [gray!30] (0,0) -- (2.5,0.5) -- (3,3) -- node[midway,sloped,above,color=black] {Nearest plane} (0.5,2.5) -- cycle;
    \node [points={purple}]  at (0.7,0.8) {};
    \node [points={purple}]  at (1,1.5) {};
    \node [points={purple}]  at (1.3,1) {};
    \node [points={purple}]  at (1.5,2) {};
    \node [points={purple},label=right:{\color{purple}$\mat q_j$}] (q) at (2,1.8) {};
    \node [points={blue},label=right:{\color{blue}$\mat p_j^w$}] (p) at (3,1) {};
    \draw [stealth-] (p) -- (q); 
    \draw [stealth-] (p) -- node[midway,below] {$\mat u_j$} (1.9,1.22) coordinate (o);
    \draw [dashed] (o) -- (q);
    \draw ($(o)!0.3!(q)$) -- ++(0.17,-0.03) -- ($(o)!0.15!(p)$);
    \node [points={purple},label=right:{: Point in map}] at (3.8,0.8) {};
    \node [points={blue},label=right:{: Point in scan}] at (3.8,0.4) {};
  \end{tikzpicture}
  \caption{The measurement model.}\label{fig:measurement model}
\end{figure}

At a certain moment, the LiDAR acquires $m$ sampling points $\{\mat p_j\}_{j=1}^m$. 
Each point $\mat p_j$ transformed into the word coordinate system is denoted
as $\mat p_j^w:=(\mat K\Gamma(\mat T))*\mat p_j$, where $*$ represents 
a group action of $\SE(3)$ on $\mathbb{R}^3$:
$(A,a)*x:=Ax+a$.
Search for the five points closest to $\mat p_j^w$ and fit a plane, 
then $\mat p_j^w$ should be in this plane (Fig. \ref{fig:measurement model}).
Therefore, the system output $h_j:\mfd M\to\mathbb{R}$ can be configured as \cite{fastlio2}
\[
  h_j(\xi)=\mat n_j^\top\bigl(\mat p_j^w-\mat q_j\bigr) ,
\]
where $\mat n_j$ is the normal vector of the plane and $\mat q_j$ is an arbitrary point in the plane.

According to \cite{EqF1}, the linearized measurement equation is
\begin{equation*}
  \mat z_j=\mat H_j\varepsilon,\ 
  \mat H_j=(dh_j)_{\hat{\xi}}\circ (d\phi_{\hat X})_{\xi^0}\circ
  (d\varphi^{-1})_{\mat 0}.
\end{equation*}
The closed-form of the matrix $\mat H_j$ is
\[
  \mat H_j=\begin{bmatrix}
    \mat O_{1\times 9} & \mat O_{1\times 9} & 
    \mat H_{j,K}
  \end{bmatrix}\in \mathbb{R}^{1\times 24},
\]
where
\[
  \mat H_{j,K}=\begin{bmatrix}
    -\mat u_j^\top \bigl(\hat B*\mat p_j^w\bigr)^\wedge
     & \mat u_j^\top 
  \end{bmatrix}\in \mathbb{R}^{1\times 6}.
\]

\subsection{Gravity Constraints on Manifolds \texorpdfstring{$\mathbb{S}^2$}{S2}}

\begin{figure}[t]
  \centering
  \begin{tikzpicture}[scale=2.8,tdplot_main_coords]
    \tdplotsetrotatedcoords{-90}{80}{90}
    \draw
      [tdplot_rotated_coords] (1,0,0) arc 
      [start angle=0,end angle=180,radius=1] 
      [tdplot_main_coords] (1,0,0) arc 
      [start angle=0,end angle=-180,radius=1] ;
    \draw [tdplot_rotated_coords,->,blue,thick] (0,0,0) -- (60:1) coordinate (g) node[midway,below right] {$\mat g_k$};
    \draw [tdplot_rotated_coords,->,blue,thick] (0,0,0) -- (75:1) coordinate (g1) node[midway,above left] {$\mat g_{k+1}$};
    \draw [tdplot_rotated_coords,red] ($(0,0,0)!0.4!(g)$) arc (60:75:0.4) node [midway,xshift=2,yshift=5] {$\theta$};
    \path [tdplot_rotated_coords] ($(0,{2*sqrt(3)/3},0)!0.3!(g)$) coordinate (p1) -- ($(0,{2*sqrt(3)/3},0)!1.7!(g)$) coordinate (p3);
    \path [tdplot_rotated_coords] ([shift={(-0.15,-0.1)}]g) coordinate (p2) -- ([shift={(0.15,0.1)}]g) coordinate (p4);
    \fill [opacity=0.3] (p1) -- (p2) -- (p3) -- (p4) -- node [sloped,opacity=1,above] {\small Tangent space} cycle ;
    \draw [tdplot_rotated_coords,dashed] ($(p2)!0.5!(p3)$) -- ($(p4)!0.5!(p1)$) ;
    \draw [tdplot_rotated_coords,dashed] ($(p1)!0.5!(p2)$) -- ($(p3)!0.5!(p4)$) ;
    \draw [tdplot_rotated_coords,->,red] (g) -- ($(p2)!0.5!(p3)$) node[below=3] (q) {$\norm{\mat x}=\theta$};
    \draw [tdplot_main_coords,->,red] (0,0,0) -- (-84:0.8) node [below=3] (o) {$\mat B_{\mat g_k}\mat x$};
    \draw [->,red] (q) to[out=-80,in=10] (o);
    \node [tdplot_screen_coords] at (-0.7,0.84) {$\mathbb{S}^2$};
    \draw [thick,dashed,opacity=0.7] (1,0,0) arc[start angle=0,end angle=180,radius=1] ;
  \end{tikzpicture}
  \caption{
    Illustration of the error state on $\mathbb{S}^2$. 
    The neighborhood at $\mat g_k\in\mathbb{S}^2$ is homeomorphic to $\mathbb{R}^2$.
    $\mat x$ is a minimal parameterization of the error between $\mat g_{k+1}$ with $\mat g_{k}$.
  }\label{fig:gravity optimization}
  \vskip-15pt
\end{figure}
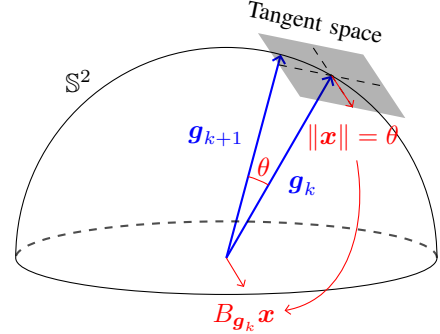

In order to maintain the minimal parameterization of gravity,
we consider gravity to be a vector with constant magnitude and direction on the sphere
$\mathbb{S}^2$ \cite{10024988}.
Since $\mathbb{S}^2$ is a two-dimensional manifold, we use a two-dimensional vector in the tangent space 
$T_{\mat g}\mathbb{S}^2\simeq\mathbb{R}^2$ as the error state for optimization,
which uses the fewest degrees of freedom to estimate gravity.

Define the direction vector of gravity as $\mat g=(x,y,z)^\top\in\mathbb{S}^2$
and $\norm{\mat g}=1$. Consider the linear map \cite{hertzberg2013integrating,10024988}
\begin{equation}
  \mat B_{\mat g}:\mathbb{R}^2\to\mathbb{R}^3\quad
  \mat B_{\mat g}=\begin{bmatrix}
    1-\frac{x^2}{1+z} & -\frac{xy}{1+z}\\[1mm]
    -\frac{xy}{1+z} & 1-\frac{y^2}{1+z}\\
    -x & -y
  \end{bmatrix}.
\end{equation}
$B_{\mat g}$ is an isometry operator, that is,
$
  \norm{\mat B_{\mat g}\mat x}=\norm{\mat x}
$
for all $\mat x\in\mathbb{R}^2$.
Through $\mat B_{\mat g}$, we can convert the vector in the tangent space 
$T_{\mat g}\mathbb{S}^2\simeq\mathbb{R}^2$ into a rotation vector, 
thereby updating the gravity.
The pseudoinverse of the matrix $\mat B_{\mat g}$ is $\mat B_{\mat g}^\top$.

Denote the direction of gravity at time $k$ by $\mat g_k\in\mathbb{S}^2$ and the error state 
of the optimal estimate by $\mat x\in\mathbb{R}^2$.
As shown in Fig.~\ref{fig:gravity optimization}, we take
\[
  \mat g_{k+1}=\exp(\mat B_{\mat g_k} \mat x)  \mat g_k\in\mathbb{S}^2.
\]
Conversely, for the two gravity directions $\mat g_k$ and $\mat g_{k+1}$, 
the error is defined as
\begin{equation}
  \varepsilon_g= \arccos(\mat g_k\cdot\mat g_{k+1})\mat  B_{\mat g_{k}}^\top (\mat g_{k}\times\mat g_{k+1}),
\end{equation}
where $\mat g_k\cdot\mat g_{k+1}$ represents the dot product of the vectors.

\section{Experiments}

In this section, we conduct a series of experiments to 
evaluate the accuracy and robustness of Eq-LIO. 
In all experiments, we compare Eq-LIO with the IEKF-based LIO 
algorithm and the EKF-based FAST-LIO2 algorithm. 
Furthermore, for fairness, all algorithms use the same 
tuning parameters in the experiments. The transformation
of the parameters between EKF and EqF is provided in the
supplementary material\cite{supp}.



\subsection{Public Datasets}

\begin{figure}[tb]
  \centering
  \includegraphics[width=\linewidth]{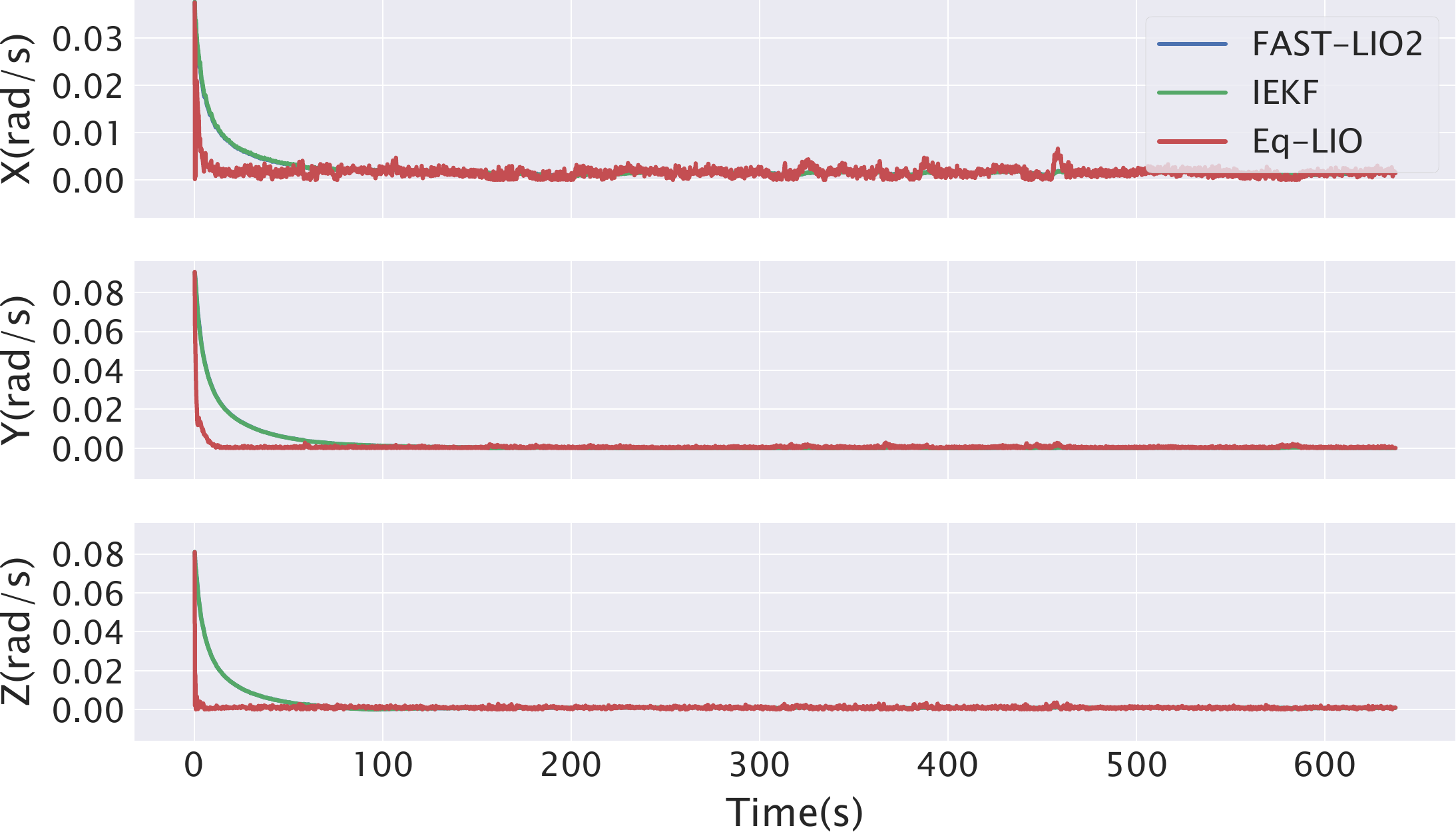}
  \vskip-5pt
  \caption{The estimated three-axis gyroscope bias sequence in the \data{Schloss-1} dataset.}
  \vskip-15pt
  \label{fig:Schloss-1 gyro bias}
\end{figure}

\textit{1) Public LiLi-OM Dataset:}
Given LiLi-OM dataset\cite{liliom} has no precise ground truth, we use
the start-end distance as an indicator to roughly estimate accuracy
of LIO. It is worth noting that the reference value is calculated by
GNSS positioning results.
The distance error for the LiLi-OM dataset is presented in Table \ref{tab:lili-om}.
The results show that the proposed Eq-LIO achieves the best accuracy 
on the three sequences, whereas the accuracy of FAST-LIO2 and IEKF 
on the \data{Schloss-2} dataset decreases significantly.
In addition, Fig. \ref{fig:Schloss-1 gyro bias} shows 
the time series of the three-axis gyroscope bias in the 
\data{Schloss-1} dataset. Interestingly,  
the proposed Eq-LIO exhibits the ability of the filter to converge 
quickly.

\begin{table}[tb]
  \centering
  \caption{Distance Errors on the LiLi-OM Dataset}\label{tab:lili-om}
  \begin{tblr}{
    colspec={*{4}{X[c,m]}},
    width=\linewidth,
    hline{1,Z}={1pt},
    hline{2},
    cell{2-Z}{1}={font=\itshape}
  }
    Error(m) & FAST-LIO2 & IEKF & Eq-LIO \\
    Schloss-1 & 0.60 & 0.57 & \textbf{0.53} \\
    Schloss-2 & 11.50 & 8.75 & \textbf{7.07} \\
    Campus-1 & 0.97 & 0.78 & \textbf{0.46} \\
  \end{tblr}
\end{table}

\begin{figure}[htb]
  \vskip-10pt
  \centering
  \includegraphics[width=\linewidth]{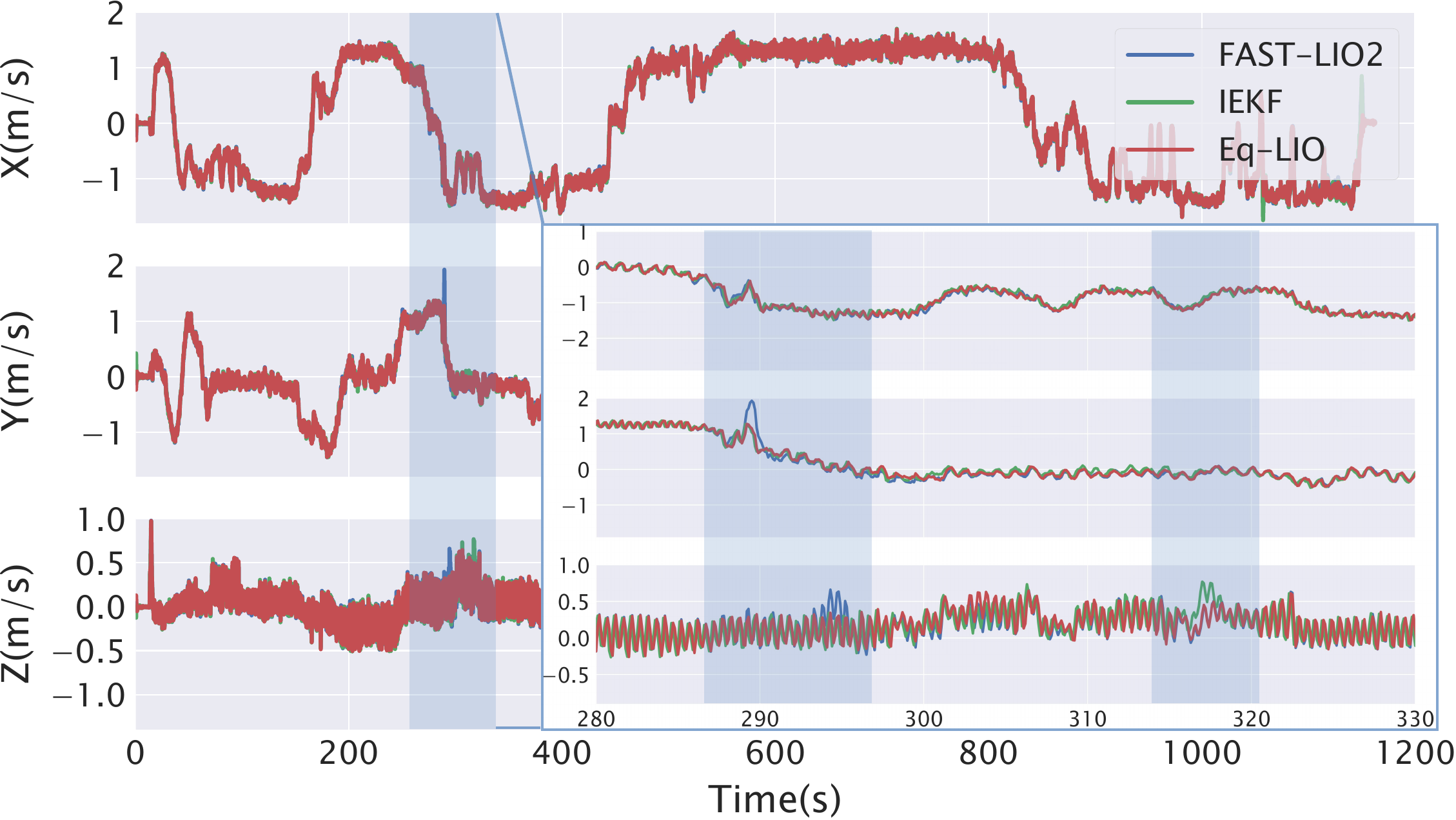}
  \vskip-5pt
  \caption{Time series plot of the estimated velocities of the three axes in the \data{hkust\_canpus\_01} dataset.}
  \vskip-5pt
  \label{fig:campus01 velocity}
\end{figure}

\textit{2) Public R3LIVE Dataset:}
In the \data{R3LIVE}\cite{R3live} dataset, we use datasets with closed 
trajectories so that we can evaluate the end-to-end error.
The end-to-end error results we obtained are shown in Table \ref{tab:R3LIVE}.
In most scenarios, Eq-LIO delivers superior results. However, 
in certain cases, such as the sequence \data{hkust\_campus\_01}, 
the scan-to-map matching strategy used by FAST-LIO2 improves 
loop accuracy but also introduces the risk of matching 
errors, which will cause large jumps in the trajectory, 
as illustrated in Fig.\ref{fig:campus01 velocity}, 
while Eq-LIO does not have this problem.
The speed curve is used here 
because it is easier to observe such jumps.


\begin{table}[htb]
  \centering
  \vskip-10pt
  \caption{End-to-End Errors on the R3LIVE Dataset}
  \label{tab:R3LIVE}
  \begin{tblr}{
    colspec={X[1.2,c,m]*{3}{X[c,m]}},
    width=\linewidth,
    hline{1,Z}={1pt},
    hline{2},
    cell{2-Z}{1}={font=\itshape}
  }
    Error(m) & FAST-LIO2 & IEKF & Eq-LIO \\
    hku\_campus\_00 & 0.17 & 0.07 & \textbf{0.06} \\
    hkust\_campus\_00 & 3.10 & 3.59 & \textbf{2.26} \\
    hkust\_campus\_01 & \textbf{0.13} & 1.51 & 0.19 \\
    hkust\_campus\_02 & 0.04 & \textbf{0.03} & 0.09 \\
    degenerate\_00 & 8.25 & 6.23 & \textbf{6.02} \\
  \end{tblr}
\end{table}

\begin{figure}[htb]
  \centering
  \includegraphics[width=\linewidth]{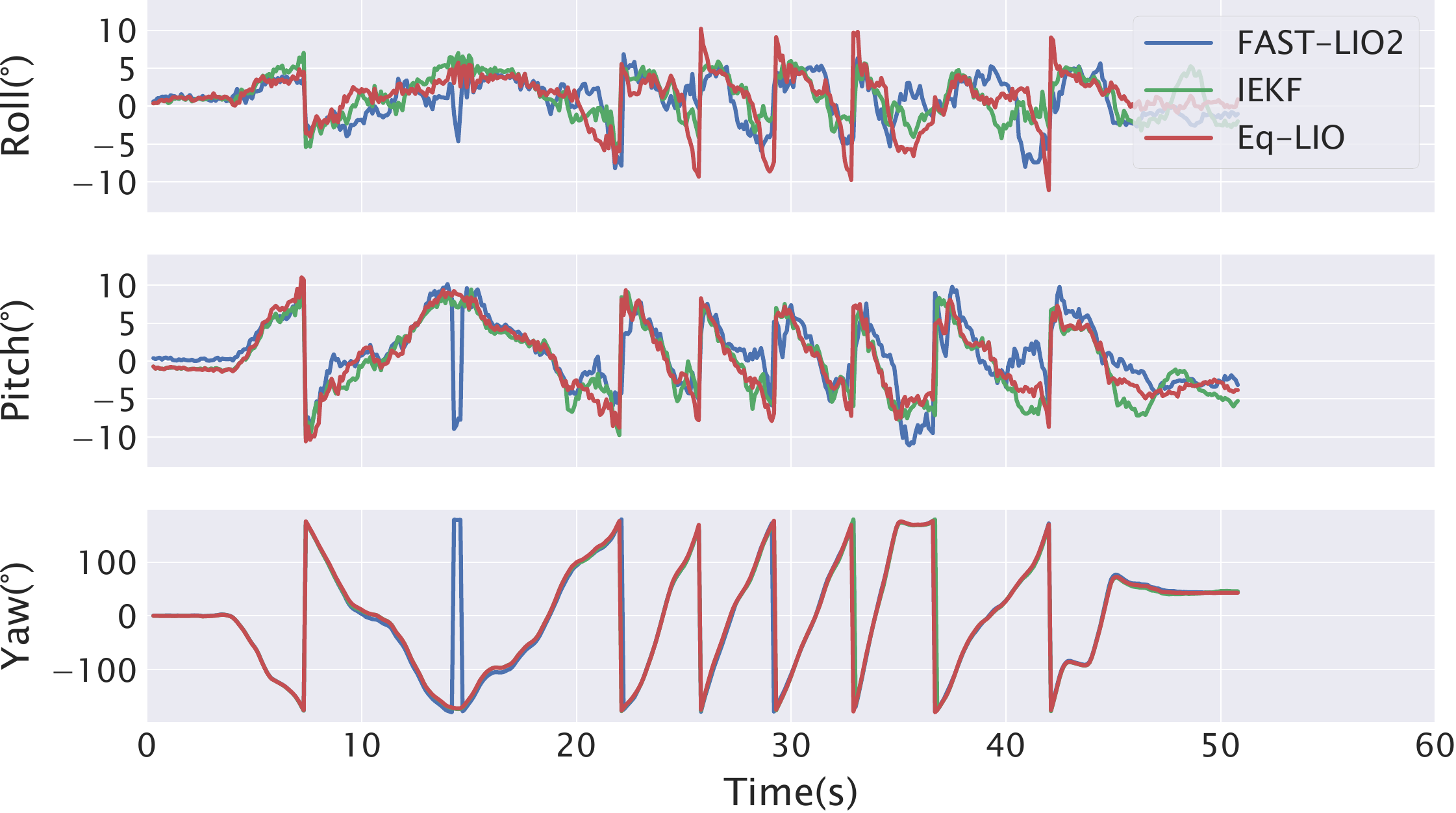}
  \caption{Estimated attitude sequences in the indoor experiment.}
  \label{fig:attitude in indoor exp}
\end{figure}


\subsection{Private Datasets}

\begin{figure}[htb]
  \centering
  \setcounter{subfigure}{0}
  \subfloat[Eq-LIO]{%
    \includegraphics[width=.33\linewidth]{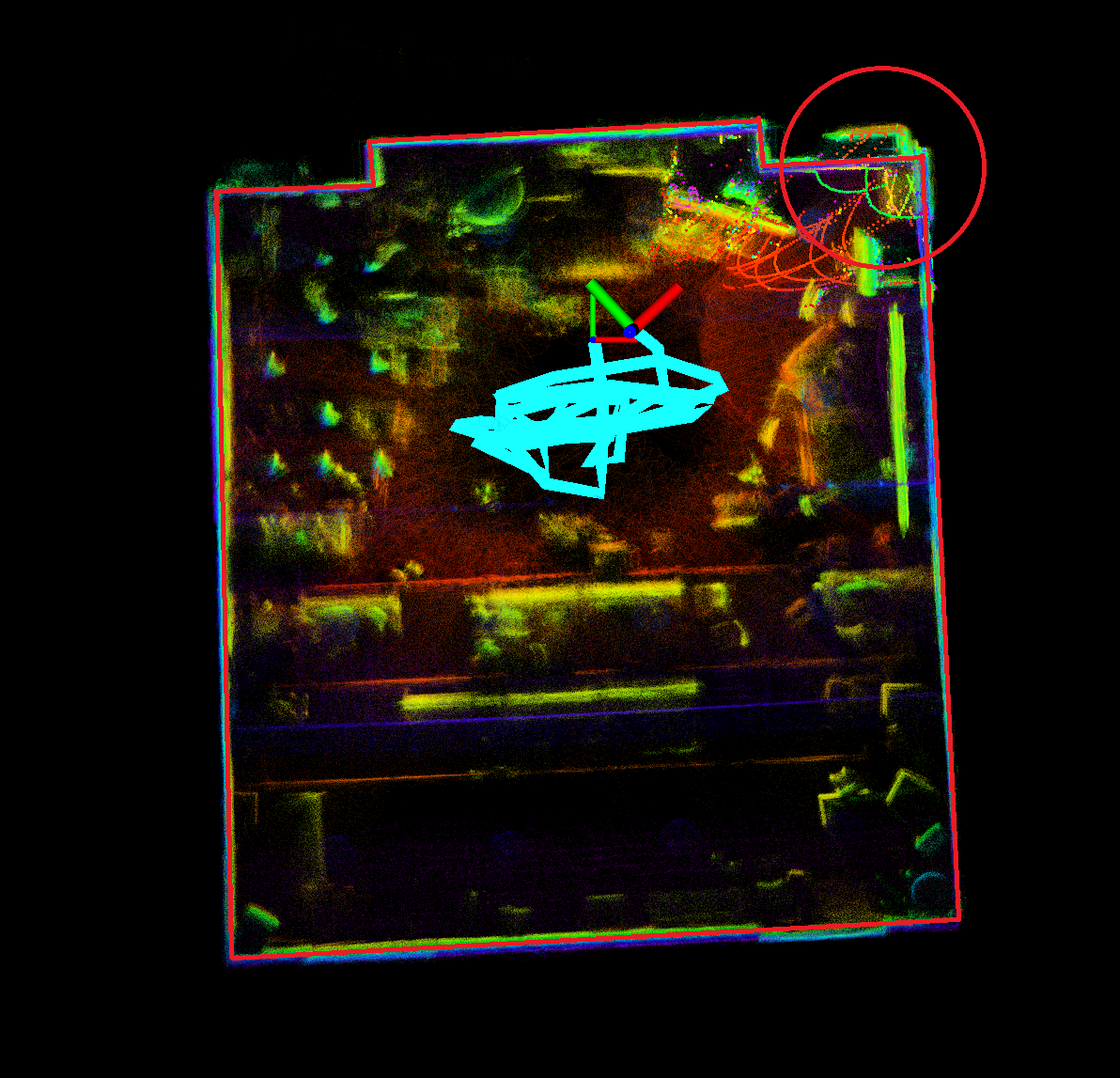}%
  }\hfill
  \subfloat[IEKF]{%
    \includegraphics[width=.33\linewidth]{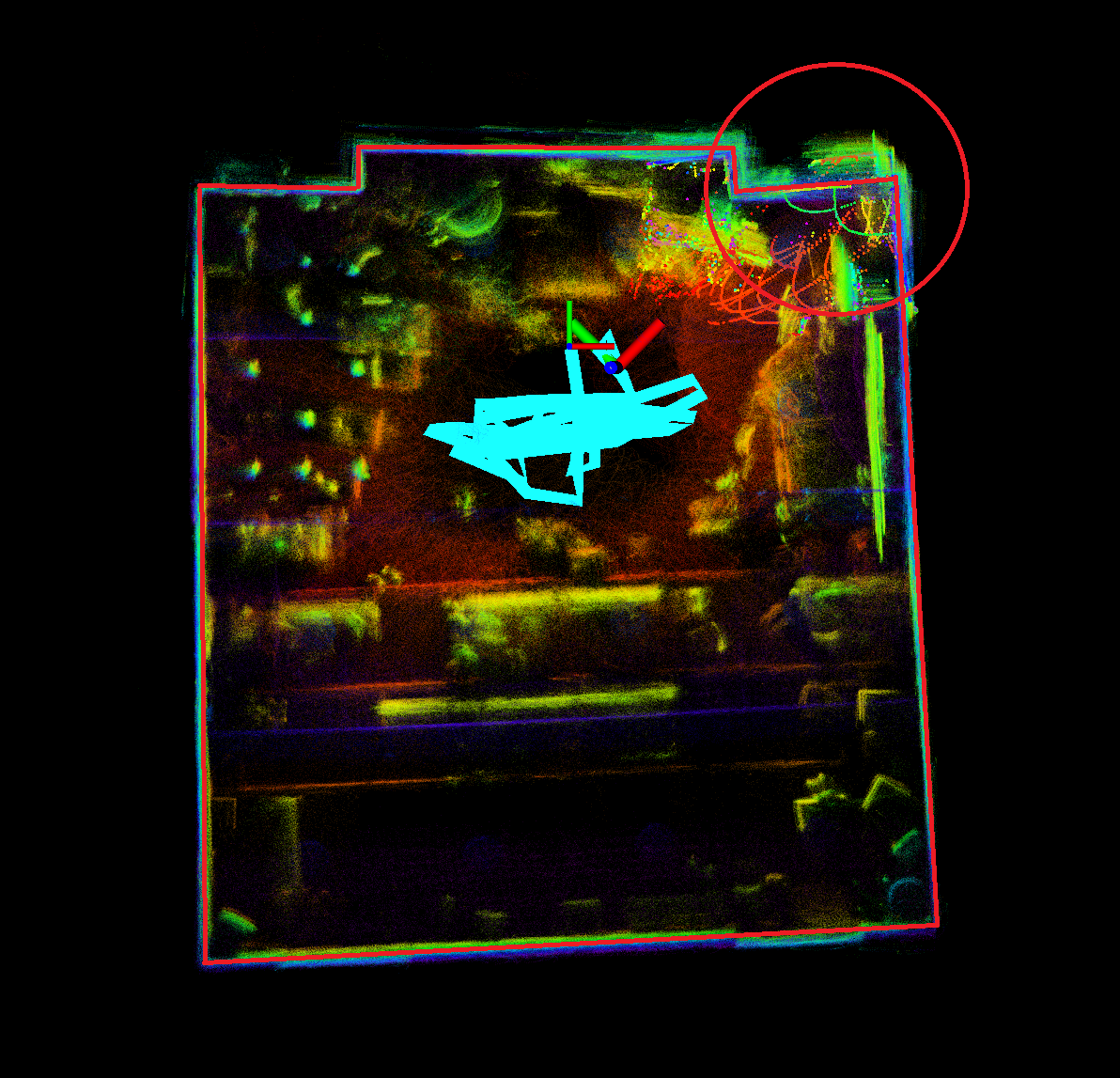}%
  }\hfill
  \subfloat[FAST-LIO2]{%
    \includegraphics[width=.33\linewidth]{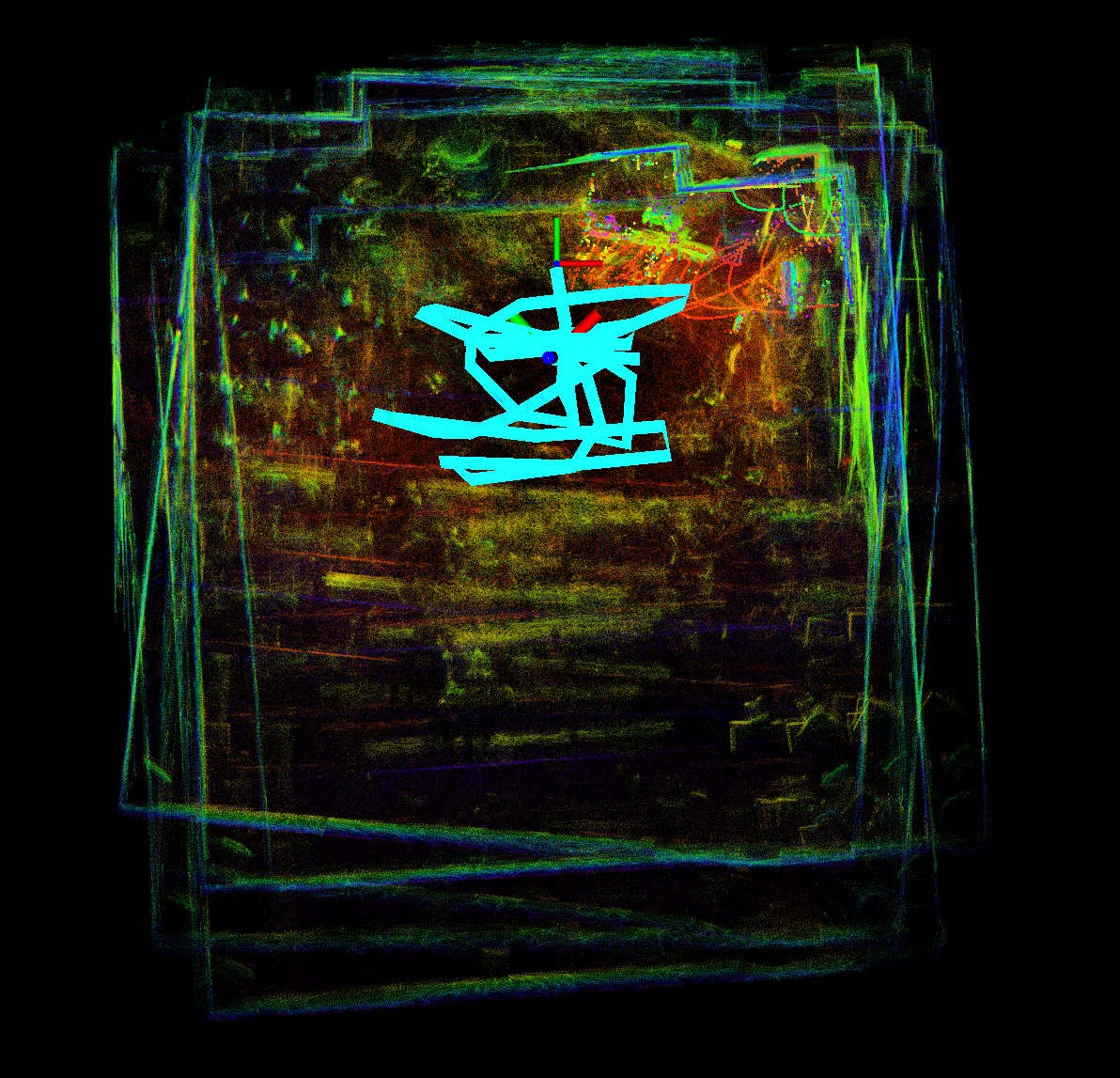}%
  }
  \caption{Mapping results of LIO in the indoor high-speed environment based on three different filtering algorithms.
  The mapping results of FAST-LIO2 for this data set are not available. The red lines depict the wall of the EqF and IEKF mapping results, which can reflect the degree of deformation of the wall.
  }
  \label{fig:map of indoor exp}
\end{figure}

\textit{1) Indoor Experiment:}
Robustness is a crucial characteristic of filter-based LiDAR-inertial odometry, 
referring to the system's ability to withstand adverse external disturbances 
such as imperfect tuning parameters, degenerate motion scenes, 
or sustained intense movement.
To induce significant changes in both rotation and velocity, the data 
collectors held the sensor in their hands and ran irregularly indoors, 
generating large and unpredictable movements. The data were 
collected using a Livox Avia LiDAR and its built-in MEMS IMU, with 
sampling frequencies of 10 Hz for LiDAR and 100 Hz for the IMU.
Fig. \ref{fig:attitude in indoor exp} 
illustrates the attitude estimation during the experiment, 
revealing dramatic changes in a short period, making it
ideal for examining the robustness of the proposed system.
The results, depicted in Fig. \ref{fig:map of indoor exp}, demonstrate 
that Eq-LIO produces the clearest mapping result and handles 
wall corners more effectively than IEKF.
The mapping results of EKF-based FAST-LIO2 exhibit significant drift, 
while Eq-LIO shows less distortion on the walls, with the upper right 
corner of the map appearing clearer. 
This suggests that when handling highly 
nonlinear motion scenarios, the model based on Eq-LIO 
exhibits a superior linearization effect, making it
more resilient to significant parameter errors and
demonstrating enhanced robustness.

\textit{2) Outdoor Experiment:}
To verify the performance of Eq-LIO in actual urban complex scenes, 
we collected data from real urban highway scenes. The dataset 
was collected on roads in Wuhan and includes 
challenging scenarios such as long straight lines and large arc U-turns.
The presence of many moving vehicles further complicates achieving robust navigation.
The hardware platform for data collection is as follows:
a GNSS receiver (Septentrio PolaRx5), GNSS
antenna (Trimble Zephyr Model2), a MEMS IMU (ADIS-16470), a tactical IMU (StarNeto XW-GI7660), and a mechanical
LiDAR (Velodyne VLP-32) were rigidly fixed on
an iron plate with roughly pre-calibrated extrinsic parameters.
We used the MEMS IMU for testing and the tactical IMU with RTK/INS to generate reference ground truth.


\begin{figure}[H]
  \centering
  \includegraphics[width=\linewidth]{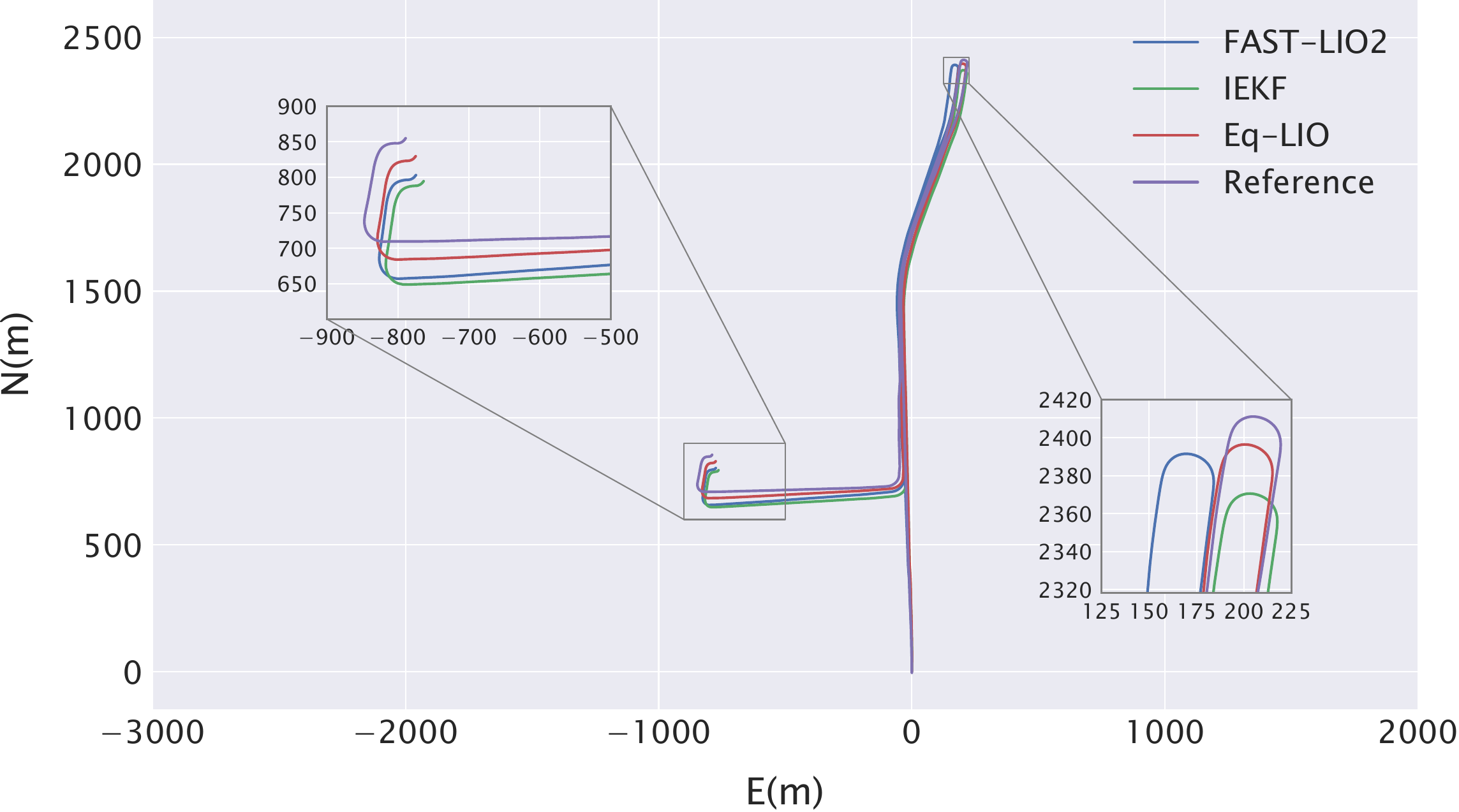}
  \caption{Trajectories estimated by three filter-based LIO algorithms in the outdoor experiment.}
  \vskip-15pt
  \label{fig:wuhan trajectory}
\end{figure}

\begin{table}[H]
  \centering
  \caption{Planar Distance Errors at the End of the Trajectory in the Outdoor Experiment.}
  \label{tab:wuhan error}
  \begin{tblr}{
    colspec={X[0.8,c,m]*{3}{X[c,m]}},
    width=\linewidth,
    hline{1,Z}={1pt},
    hline{2}
  }
      & E(m) & N(m) & Plane(m) \\
  FAST-LIO2 & 14.309 & 51.725 & 53.668 \\
  IEKF & 25.079 & 60.302 & 65.309 \\
  Eq-LIO & \textbf{13.899} & \textbf{25.124} & \textbf{28.713} \\
  \end{tblr}
  \vskip-5pt
\end{table}

As shown in Fig. \ref{fig:wuhan trajectory}, all algorithms exhibit some 
errors due to the long trajectory and the presence of 
numerous moving objects. However, at both the U-turn and the end of 
the trajectory, the proposed Eq-LIO demonstrates superior performance, 
accurately estimating the scale of trajectory. 
Table \ref{tab:wuhan error} shows the planar distance errors 
of different algorithms at the final moment. The trajectory 
length is approximately 5132 meters, with Eq-LIO exhibiting  
the smallest drift in both the east and north directions. 
The vehicle primarily traveled along the north-south axis, 
resulting in more significant inconsistency in the northward 
direction. The results indicate that Eq-LIO produces 
smaller errors under erroneous observations compared to 
IEKF and EKF, demonstrating superior consistency in the northward direction.

\section{Conclusion}

This article presents Eq-LIO.
To address the inconsistency and 
linearization point issues of traditional filtering algorithms, 
this work develops an equivariant filter for LIO systems, 
derives a new error definition based on the symmetry of the 
semi-direct product group, and incorporates gravity constraints. 
The algorithm achieves improved
accuracy and robustness without increasing computational resources. 
Experiments on various public and private datasets 
show that, compared to existing methods, Eq-LIO can achieve higher accuracy and more importantly, 
robustness to challenging motion scenes at comparable speeds.
In future, the gravity state is supposed to be enbraced into the
equivariant system.



\bibliographystyle{IEEEtran}
\bibliography{ref}

\clearpage

\includepdf[pages=-]{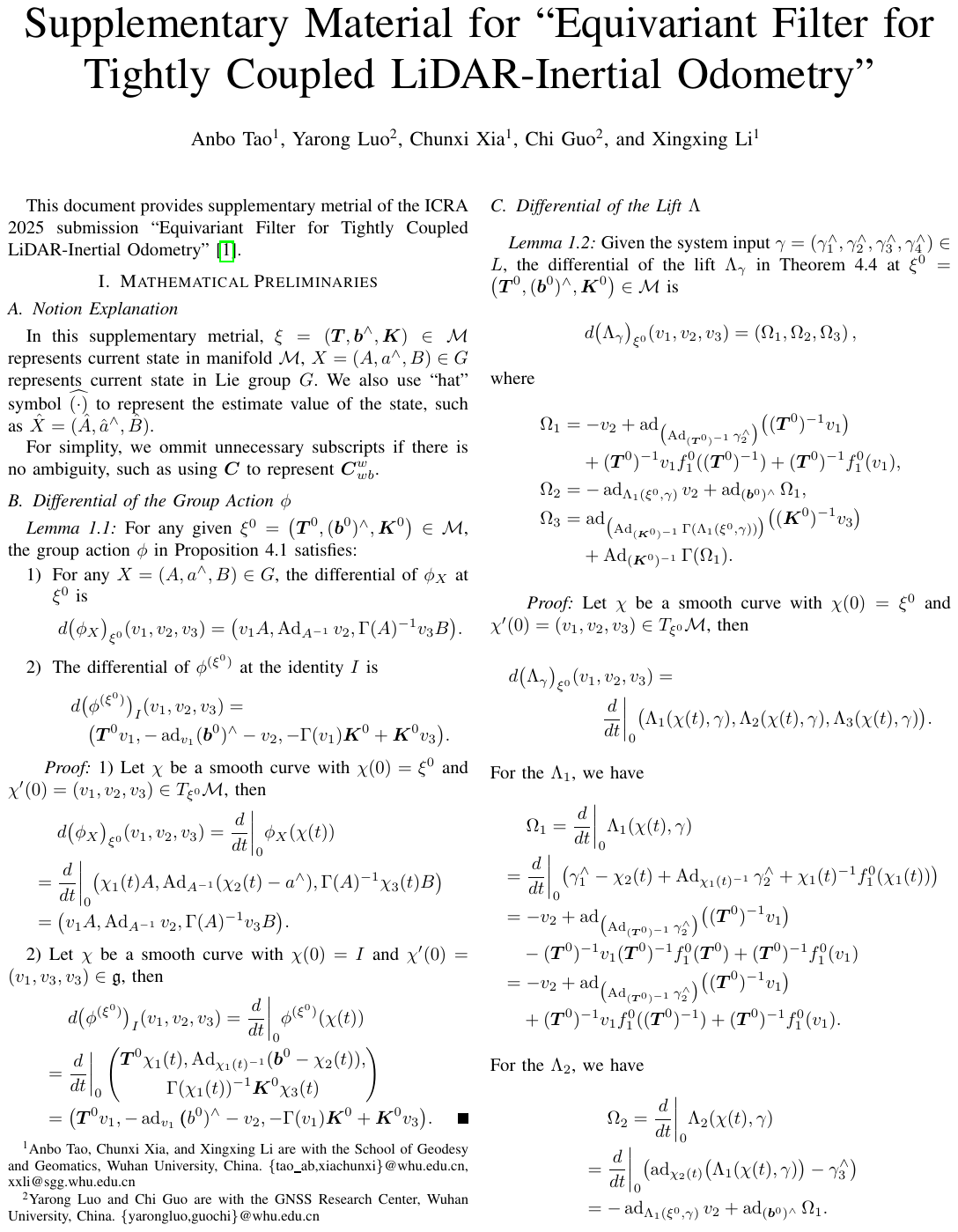}

\end{document}